\documentclass[10pt,twocolumn]{article}
\usepackage[utf8]{inputenc}
\usepackage{cvpr}
\usepackage{times}
\usepackage{epsfig}
\usepackage{graphicx}
\usepackage{amsmath}
\usepackage{amssymb}
\usepackage{amsthm}
\usepackage{algorithm}
\usepackage{xcolor}
\usepackage{subcaption}
\usepackage{blindtext}

\DeclareMathOperator*{\argmax}{\arg\,\max}
\RequirePackage{algorithmic}

\graphicspath{{fig/}{fig/dweight/}{images/}{./}} 
\usepackage[bottom]{footmisc}

\usepackage[pagebackref=true,breaklinks=true,letterpaper=true,colorlinks,bookmarks=false]{hyperref}

\cvprfinalcopy 

\def\cvprPaperID{3474} 

\ifcvprfinal\fi
\begin{document}

\title{Diagnostic Visualization for Deep Neural Networks Using Stochastic Gradient Langevin Dynamics}

\author{Biye Jiang\footnotemark[1] \hspace{8pt} David M. Chan\footnotemark[1] \hspace{8pt} Tianhao Zhang \hspace{8pt} John F. Canny 
\\
University of California, Berkeley \\ 
\tt{\small{<bjiang,davidchan,bryanzhang,canny>@berkeley.edu}}
}
\maketitle

\begin{abstract}
    The internal states of most deep neural networks are difficult to interpret, which makes diagnosis and debugging during training challenging. Activation maximization methods are widely used, but lead to multiple optima and are hard to interpret (appear noise-like) for complex neurons. Image-based methods use maximally-activating image regions which are easier to interpret, but do not provide pixel-level insight into why the neuron responds to them. In this work we introduce an MCMC method: Langevin Dynamics Activation Maximization (LDAM), which is designed for diagnostic visualization. LDAM provides two affordances in combination: the ability to explore the set of maximally activating pre-images, and the ability to trade-off interpretability and pixel-level accuracy using a GAN-style discriminator as a regularizer. We present case studies on MNIST, CIFAR and ImageNet datasets exploring these trade-offs. Finally we show that diagnostic visualization using LDAM leads to a novel insight into the parameter averaging method for deep net training. 
\end{abstract}

\let\thefootnote\relax\footnote{$^*$ Denotes equal contribution}


\section{Introduction}

\begin{figure}
\centering
\includegraphics[width=\linewidth]{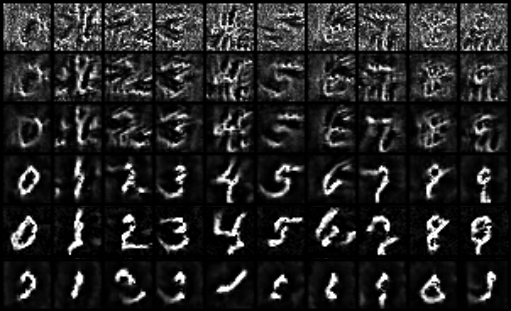}

\vspace{5mm}
\includegraphics[width=\linewidth]{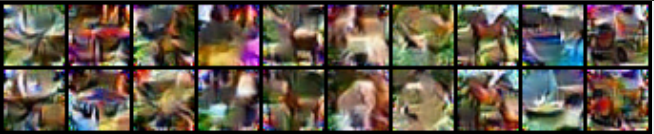}
\caption{\textbf{Top:} Example maximally-activating output-layer visualizations from a LeNet trained on MNIST. The columns represent visualizations of different output neurons The rows represent different possible image manifolds corresponding to differing amounts of regularization by an adversarial discriminator. \textbf{Bottom:} Examples visualizations from CIFAR-10 showing the diversity of potential image generations.The rows show visual diversity, while the columns represent different class output-neuron responses. }
\label{fig:headline}
\end{figure}

Deep neural networks (DNNs) have seen wide adoption, but their ability to learn ad-hoc features typically makes them hard to understand and diagnose. Visualization for deep networks has a long history with activation maximization approaches  \cite{springenberg2014striving, smilkov2017smoothgrad,nguyen2016multifaceted,nguyen2016synthesizing,olah2017feature}. These approaches have produced impressive visualizations but ad-hoc regularization is needed to produce images that are interpretable (not noise-like). This phenomenon is related to adversarial fooling images \cite{szegedy2013intriguing}. These perturbations maximally activate a (false) label class and are noise-like and imperceptible.

Image-based and salience methods \cite{DBLP:journals/corr/SelvarajuDVCPB16} explore neuron activations on specific images to determine whether a given label is ``right for the right reason'' \cite{DBLP:journals/corr/abs-1803-09797}. But they do not allow for systematic examination of the set of all highly-activating image patches. 

This paper aims to combine the affordances of image-based and pixel-based approaches for diagnostic visualization: for diagnosis we don't care about pathological (non-physical) activation patterns, rather we want the maximally activating patterns that lie in the manifold of real images. 
We therefore use a GAN-like discriminator as a regularizer: the discriminator is trained to distinguish real images from the visualizations generated by our MCMC method. The discriminator output ``false'' is then used as a regularizer which we minimize during MCMC sampling. Since a strong discriminator weight tends to cause mode-locking (sampling only within one label class), we reduce the weight of the discriminator during exploration. We also adjust sampler temperature allowing us to fully explore the space of highly-activating images. 

The simplest deep net visualization method is activation maximization (AM) \cite{erhan2009visualizing}, which generates (artificial) images which maximally activate a given neuron.  AM can be posed as an optimization problem over the pixel matrix space $\Pi = \Bbb{R}^{w\times h}$: Given $f_\alpha(x)$, the activation of a neuron $\alpha$ in a network $f$ on the input image $x$, find a maximally-activating image $x \in \Pi$.
This process is highly under-constrained however, so a regularization term $R(x)$ is added to produce a unique and hopefully interpretable image. Olah \etal \cite{olah2017feature} summarized several different regularization techniques which can make results smoother \cite{mahendran2016visualizing} or more human-interpretable \cite{nguyen2016synthesizing}. The regularizer yields the following optimization problem:
\begin{equation}
\label{eq:am}
\argmax_{x|x \in \Pi} \quad f_\alpha(x) + \lambda R(x)
\end{equation}

A challenge with this approach  \cite{mahendran2016visualizing,nguyen2016multifaceted,nguyen2016synthesizing} is that ad-hoc regularizers (smoothness etc) work well for simple activation patterns,
but not the complex patterns seen in later-layer neurons. Such ``deep dream'' visualizations are appealing but may be far from the naturalistic images that the neuron most strongly responds to. 

Secondly the AM optimization problem is usually non-convex: there are many exemplars of a complex class like ``car'', or even its lower-level features. This was observed by Nguyen \etal \cite{nguyen2016multifaceted}, who found that single neurons can have multi-faceted behavior, in that they respond to many different stimuli. The paper \cite{erhan2009visualizing} explored the space of activating images using different initializations, but unfortunately there is no guarantee that such an approach will cover this space. Our MCMC does provide such guarantees, although without precise time bounds. In practice by annealing temperature, we find that thorough exploration is possible. 

More recently, generative adversarial networks (GANs) \cite{zhou2017activation, nguyen2016synthesizing, zhou2018activation} have been used to improve the realism of AM images. For example, Nguyen \etal \cite{nguyen2016synthesizing} use a "Deep Generator Network" for this purpose. A
difficulty with using discriminator loss directly is mode-locking: images will typically fall into discrete modes (subclass) that the generator cannot move between. Zhou \etal \cite{zhou2018interpreting} attempt to address this issue by linking human-defined semantic concepts with feature activations, however a human selection of semantic concepts can skew the users understanding of how the network is learning. 

Here we define \textbf{Langevin Dynamics Activation Maximization (LDAM)}. LDAM systematically samples images from the  based on the activation of a given neuron. LDAM is a gradient-based MCMC (Monte Carlo Markov Chain) algorithm which uses Stochastic Gradient Langevin Dynamics \cite{welling2011bayesian} to sample directly from the distribution:
\begin{equation}
\forall x \in \Pi \quad P(x) \propto \exp \left( \frac{f_\alpha(x) + \sum_i \lambda_i R_i(x)}{T} \right)
\end{equation}
where $T$ is a temperature parameter. This distribution is motivated in Section \ref{sec:motiv}. Using LDAM, we can  traverse the image manifold using a live animation while allowing users to manipulate the hyper-parameters of the sampler. Raising temperature or reducing
regularization lead to noisy, non-physical ``intermediate'' images that collapse to different naturalistic images when the original values of those parameters are restored. Smilkov \etal  \cite{smilkov2017direct} show that such direct manipulation can be particularly beneficial for users' understanding of a model, particularly during the training phase.


\begin{figure}
\centering
\includegraphics[width=\linewidth]{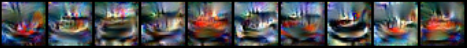}
\caption{Monte Carlo sampling can obtain samples from the entire space. For example, the above set of boats were all generated using LDAM at run-time in CIFAR-10 during the same run. }
\label{fig:mcmc}
\end{figure}
Finally, adjustment of the regularization parameter allows the designer to morph between a neuron's ``true'' (unbiased by the image manifold) and image-biased activation patterns, which in itself can 
be useful for diagnosis. 
The main contributions of the paper are: 

\begin{enumerate}
\item We introduce a sampling algorithm, LDAM, based on Stochastic Gradient Langevin Dynamics to explore the activating-image manifold. (Section \ref{sec:methods})
\item We discuss two methods of regularization: L2 and discriminator-based. We show that L2 regularization has
a simple, but not widely-recognized interpretation. (Section \ref{sec:regularization} ) 
\item We evaluate LDAM using case-studies of several image datasets and model architectures. We present a novel insight into the benefits of model parameter averaging from LDAM visualization. (Sections \ref{sec:mnist}, \ref{sec:cifar},\ref{sec:ImageNet})
\end{enumerate}

\section{Related work}
\subsection{Activation maximization}

 Olah \etal summarizes different activation maximization methods in \cite{olah2017feature} and describe several applications in \cite{olah2018the}. 
To produce cleaner, more interpretable images,  various regularizers have been proposed for AM. Mahendran \etal \cite{mahendran2016visualizing} discussed total-variation (TV) regularization and jitter. Total-variation regularization reduces inter-pixel variation, while Jitter regularization helps  create sharper and more vivid reconstructions. 

Nguyen \etal \cite{nguyen2016multifaceted} explored the multimodality of AM. By using images from the training set clustered according to activation as initializers, \cite{nguyen2016multifaceted} created a variety of images, exposing the diversity of an image class. In a follow-up, \cite{nguyen2016synthesizing} uses a pre-trained generator to produce realistic images corresponding to high-level neurons. Such images are interpretable but is only accurate for high-level layers and fully-trained networks. In contrast, LDAM uses an MCMC sampler to directly ``invert'' the function learned by a given neuron and to visualize  it. This approach is therefore not limited to output neurons, or fully-trained neurons and is more versatile for diagnosis. Finally following \cite{nguyen2016synthesizing} we use an adjustable adversarial discriminator. The discriminator biases the MCMC preimages toward interpretability without an explicit 

Guided back-propagation \cite{springenberg2014striving} is another gradient-based method that is used to visualize the saliency map for a given input image. Smilkov \etal \cite{smilkov2017smoothgrad} improves the results from  \cite{springenberg2014striving} by adding gaussian perturbations to gradients multiple times and average the results to achieve smoother gradients. LDAM borrows the idea from \cite{smilkov2017smoothgrad} by performing sample-averaging on a series of noisy samples to improve the interpretability of the results.

\subsection{Stochastic Gradient Langevin Dynamics}

One of the biggest problems with activation maximization techniques is that given an initialization, there is only a single maximum which can be achieved by gradient ascent. Erhan \etal \cite{erhan2009visualizing} also found that even with relatively distributed initialization, activation maximization produces only very few unique samples in practice.  This is particularly unsatisfying for visualizing images which maximize a neuron activation, because we would like to see a large number of unique images corresponding to a neuron's activation (or a cluster of neuron's activations). 

Thus, to sample from this space LDAM borrows the optimization technique from Welling \etal \cite{welling2011bayesian}, which uses a combination of a stochastic optimization algorithm with Langevin Dynamics which injects noise into the parameter updates in such a way that the trajectory of the parameters will converge to the full posterior distribution rather than just the maximum a posteriori mode. 

A few optimizations over \cite{welling2011bayesian} have been studied. Feng \etal \cite{feng2017learning} uses Stein Variational Gradient Descent to improve the diversity of the samples drew from a posterior distribution. Neelakantan \etal  \cite{neelakantan2015adding} use gradient noise to help train deep neural nets, and Gulcehre \etal \cite{pmlr-v48-gulcehre16} propose using noisy activation function to allow the optimization procedure to explore the boundary between the degenerate (saturating) and the well-behaved parts of the activation function.

\section{Proposed Algorithm}
\label{sec:methods}

\subsection{Motivation}
\label{sec:motiv}

We introduce the notation we will use throughout this section. Let $f_\alpha(x; \theta)$ be the activation of a neuron $\alpha$ in a neural network $f$ with parameters $\theta$ given the image $x$ from some pixel space $\Pi$ represented as $\Bbb{R}^{w\times h}$. We also let $\Xi$ be the subset of $\Pi$ representing real images. 

The goal of our algorithm (like all AM algorithms) is to sample images $x \in \Xi$ with high values $f_\alpha(x)$. To do so, we first define the random variable $X$, and the probability mass function $P(X = x | f_\alpha, \theta)$ as:
\begin{equation}
\label{eq:prelim}
\forall x \in \Xi \quad P(X = x | f_\alpha, \theta) \propto \exp \left( \frac{f_\alpha(x)}{T} \right)
\end{equation}
As we can see, the probability of sampling any image from the manifold is proportional to the activation of that image by the classifier $f_\alpha(x;\theta)$. The only issue with the formulation in Equation \ref{eq:prelim} is that it depends on an image manifold $\Xi$, which may not be available for direct sampling, or may be difficult to sample from. To resolve this issue, we can sample directly from pixel space $\Pi$, giving rise to the PMF:
\begin{equation}
\label{eq:pixel_space_prelim}
\forall x \in \Pi \quad P(X = x | f_\alpha, \theta) \propto \exp \left( \frac{f_\alpha(x)}{T} \right)
\end{equation}
The issue with equation \ref{eq:pixel_space_prelim} is that we are now no longer constrained to the image manifold $\Xi$, but a much more general space $\Pi$. To rectify this we use a regularization function $R(x)$ to define a second PMF over the images of $\Pi$:
\begin{equation}
\label{eq:suitability}
\forall x \in \Pi \quad P(x \in \Xi) \propto \exp \left( \frac{R(x)}{T} \right)
\end{equation}
We make the assumption that $P(x \in \Xi)$ and the probability that $x$ activates $f_\alpha(x)$ are independent to give us our final PMF, $h_X(x)$ for $X$ from which we can sample:
\begin{equation}
\label{eq:final_eq}
\begin{split}
\forall x \in \Pi \quad h_X(x) & = P(X = x) \\
& = P(x | f_\alpha, \theta)P(x \in \Xi) \\
& \propto \exp \left(\frac{f_\alpha(x) + \lambda R(x)}{T} \right)
\end{split}
\end{equation}

This factorization into activation and regularization parts means that at the time of diagnosis, the user can smoothly vary the regularization function to view the effects that different image-manifold assumptions have on the sampling process independent of the classifier (something which is impossible for traditional activation maximization techniques). The parameter $\lambda$ in equation \ref{eq:final_eq} is also important, as it represents a trade-off between sampling images that lie on $\Xi$, and images that are highly activating the selected neuron. 


\begin{figure}
    \centering
    \includegraphics[width=\linewidth]{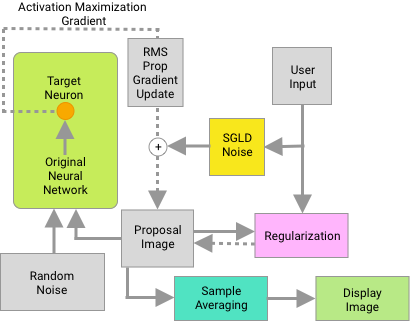}
    \caption{Diagram of the proposed system. Dashed lines represent gradient, while solid lines represent input images/control inputs. We first initialize the proposal with random noise, then update it, computing the activation maximization gradient and adding white noise. We then regularize the proposal based on the specified regularization, and update the display image. This process is then repeated}
    \label{fig:system_design}
    \vspace{-5mm}
\end{figure}

\subsection{Activation Maximization with Stochastic Gradient Langevin Dynamics (LDAM)}

A visual overview of our algorithm is given in Figure \ref{fig:system_design}. In order to sample from the distribution proposed in the previous section, we propose LDAM, an algorithm which relies on Stochastic Gradient Langevin Dynamics (SGLD, a form of Monte Carlo sampling) rather than pure gradient ascent. While pure gradient ascent could be used to optimize Equation \ref{eq:am}, it will find only a nearby local optimum. Langevin dynamic sampling will explore the entire distribution given enough time. Raising and lowering temperature (e.g. by controlling the added noise) accelerates the exploration. 

In Bayesian learning, SGLD is traditionally used to generate samples from the posterior distribution $P(\theta | D)$ where $\theta$ is the model parameters, and $D$ is the training data \cite{welling2011bayesian}. We flip the traditional sense, and sample from $\Pi$ according to the PMF given in equation \ref{eq:final_eq}. LDAM generates a sample $x_t$ at time $t$ from the distribution of $X \sim h_X(x)$ by injecting suitably scaled Gaussian noise in the gradient direction:
\begin{equation}
\label{eq:ldam}
\begin{split}
x_t & = x_{t-1} + \beta_t \Delta x_{t-1} \\
\Delta x & =   \nabla_x f_\alpha(x) + \sum_i \lambda_i \nabla_x R_i(x) + \eta \quad \eta \sim N(0, \sigma)
\end{split}
\end{equation}

In this case $\lambda$, $\beta$ and $\sigma$ are hyper-parameters which control the sampling process. In LDAM we sample our initial $x_0$ using isotropic Gaussian noise. Typically step sizes are $\beta_t = a(b + t)^{-\mu}$ decaying polynomially with $\mu \in (0.5, 1]$. In our implementation, we leave the step size up to the user as a trade-off between local and global exploration.

It is worth noting that for some regularizers (for example the discriminator), the scale of $\nabla_x f_\alpha(x)$ and $\nabla_x R(x)$ could be extremely different. Thus instead of directly using the gradient, we use a normalized gradient for those regularizers. The detailed algorithm can be seen in Algorithm \ref{alg:langevin}. 

\begin{algorithm}
   \caption{Langevin Dynamics Activation Maximization (LDAM)}
   \label{alg:langevin}
\begin{algorithmic}[1]
   \STATE {\bfseries Input:} Network $f_\alpha(x)$, Regularization functions $R_i(x)$
   \STATE {\bfseries Initialize:} Random Image $x_0$, $v_0 = 0$
   \FOR{t = 1,2,3...} 
   \STATE Forward pass: Compute $f_\alpha(x_t), R_i(x_t)$.
   \STATE Backward pass: Compute $\nabla_x f(x_t), \nabla_x R_i(x_t)$.
   \STATE Get normalized gradient $\overline{\nabla_x f_\alpha(x_t)}$ and $\overline{\nabla_x R_i(x_t)}$
   \STATE Draw $\eta_{t}$ from $N(0,\sigma_t)$
   \STATE $g_t = \overline{\nabla_x f_\alpha(x_t)} + \sum_i \lambda_i \overline{\nabla_x R_i(x_t)} +  \eta_t$
   \STATE $v_{t+1} =  \gamma v_t + \beta_t g_t$
   \STATE $x_{t+1} = x_t +  v_{t+1}$
   \STATE Process new sample $x_{t+1}$
   \ENDFOR
\end{algorithmic}
\end{algorithm}

After obtaining each new sample $x_{t+1}$, we could either directly visualize it as a live-animation in the interface, or compute the moving average of all the samples received so far. In Sections \ref{sec:mnist}, \ref{sec:cifar}, \ref{sec:ImageNet}, we will show that simple sample averaging can greatly reduce noise and improve the interpretability of the generated images. Like many MCMC style algorithms, LDAM requires a burn-in period. While the burn-in period can be determined directly as discussed in \cite{welling2011bayesian}, the computation is not reasonably efficient. Thus, to approximate the exact calculation, we wait for activation to reach a certain threshold and become stable.
\vspace{-3mm}
\section{Regularization}
\label{sec:regularization}
Choosing the right regularization function (Equation \ref{eq:suitability}) is extremely important. If we introduce regularization without realizing the effects that it has on the prior distribution, then we may distort the diagnostic power of our algorithm.

\subsection{L2 Regularization}

The first regularization function that we consider is the $L2$ norm of the generated image (treating the image as a vector in $\Bbb{R}^{wh}$), i.e $R(x) = -\frac{1}{2}||x||^2$. While L2 has been frequently used in previous work to ``clean up'' generated activation maps, its role in pixel interpretability does not seem to have been noted. Namely that at a local optimum of activation, the \textit{activation is a multiple of the gradient}.
\begin{equation}
\begin{split}
x^*  = \argmax_{x \in \Pi} f_\alpha(x) & - \lambda\frac{1}{2}||x||^2  \implies \\
\nabla_x f_\alpha(x^*) -\lambda x^* & = 0 \implies \\
\nabla_x f_\alpha(x^*) & =\lambda x^* 
\end{split}  
\end{equation}
An important corollary of this observation is that for neurons whose output is linear in the image values {\em the L2-regularized activation equals the filter weights}. i.e. the L2-regularized AM reproduces the filter weights in the first convolutional layer, and generalizes in a natural way to gradients in other layers. The filter weights from the first
layer provide a simple validation for the
LDAM method, which should compute the same values for first-layer neurons. 

\subsection{Discriminator-Based Regularization}
\label{sec:discriminator}

Our goal in this work is to maximize the interpretability of AM maps from both pixel and image perspectives. A variety of subjective regularizers have been used in prior work, but each distorts the pixel interpretability of the basic L2-regularized model. In addition, we often want to use our visualization methods during the training process, and the artifacts that may hinder image-perspective interpretability will change over time. So we seek to apply a regularization function which maximizes image interpretability at each stage in training with minimal pixel-level distortion. The solution is to train a discriminator as used in Generative Adversarial Networks, and use the discriminator gradient to improve AM interpretability. By training a network to distinguish between real images $x_R \in \Xi$ and fake images $x_F \in \overline{\Xi}$ we are implicitly constructing a probability distribution $P(x \in \Xi) \propto D(x; \theta')$, where $D$ is a discriminator, and $\theta$ are the weights of the discriminator. Thus, we can take for our regularization function $R(x) = D(x; \theta')$. By periodically re-training the discriminator, we ensure that it tracks and minimizes the image artifacts that AM produces at various stages of training. 


While the discriminator could take many shapes and forms, for simplicity in our experiments we attempt to use a structure which is as similar to the original classification networks as possible - they all have the same input size and similar hidden layer structure. Algorithm \ref{alg:adversarial} gives an outline of our algorithm using the online-trained discriminator for regularization.




\begin{algorithm}[h]
  \caption{LDAM with a Discriminator}
  \label{alg:adversarial}
\begin{algorithmic}[1]
  \STATE {\bfseries Input:} Network $f_\alpha(x; \theta)$, discriminator $D(x; \theta')$
  \REPEAT
  \STATE {\bfseries Initialize:} Random image $x_0$, $t = 0$
  \REPEAT
  \STATE Generate next sample $x_{t+1}$ using LDAM with $f_\alpha(x; \theta)$ and $R(x) = D(x; \theta')$
  \STATE $t = t + 1$
  \UNTIL{$D(x_t) > 0.9$}
  \STATE Collect half batch of generated samples with label 0, and half batch of real images with label 1
  \STATE Update the discriminator weights $\theta'$ using this batch.
  \UNTIL{Stop by user}
\end{algorithmic}
\end{algorithm}

\section{Case study: MNIST dataset}
\label{sec:mnist}
To show the applicability of our method to the diagnosis of neural architectures, we perform case studies on the MNIST, CIFAR-10, and ImageNet datasets. Through these experiments we explore how LDAM functions in practice, and how we can vary the parameters of the image manifold to achieve clearly understandable results. In addition, we explore the choice of regularization function $R(x)$, and explore how influencing the image manifold can provide interesting insights into the actions of neurons in a classifier. 

We first apply our system to the classic LeNet model trained on the classic MNIST handwritten digits dataset \cite{lecun1998gradient}. The MNIST dataset contains 60000 training images and 10000 testing images. The images are $28\times 28$ in gray scale The network we use is similar to the original LeNet-5 \cite{lecun1998gradient}. It contains 2 convolution layers with $5 \times 5$ kernel, followed by 3 fully connected layers. We train the network until convergence using RMSProp with momentum. 

\subsection{Output Neurons}
We begin by exploring the output-layer neurons. By running LDAM on these neurons, we should get images which correspond to human-labeled classes. Thus, these activation inputs should be easily interpretable. In this portion of the diagnosis, we are examining neurons which lie before the final softmax layer as the optimization targets. As mentioned in \cite{olah2017feature}, using neurons in this layer can generate more human-readable images compared to neurons after the softmax; target neurons after the softmax will prefer image patches that are unique to a given class, possibly removing features that are relevant to other classes. 

\begin{figure}[t]
\begin{subfigure}[t]{\linewidth}
\includegraphics[width=\linewidth]{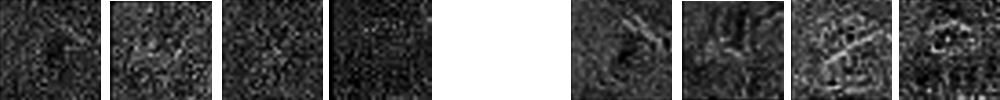} 
\caption{LDAM samples from pre-softmax neurons in the 0, 4, 8 and 9 classes. Left: No regularization. Right: L2 regularization.}
\label{fig:am_mnist_pure}
\end{subfigure}
\hfill 
\vspace{2mm}
\begin{subfigure}[t]{\linewidth}
\includegraphics[width=\linewidth]{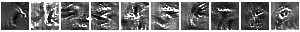} 
\caption{MNIST samples from all 10 classes using both L2 regularization and sample averaging.}
\label{fig:am_mnist_avg}
\end{subfigure}
\hfill
\centering
\caption{The effect of L2 regularization  and sample averaging on the sample space with LDAM in MNIST}
\label{fig:am_mnist}
\vspace{-9pt}
\end{figure}

Following the steps in Algorithm \ref{alg:langevin}, we start the sampling procedure from random images, and then use langevin dynamics to create proposals for all 10 classes. In order to visualize neurons with both negative and positive correlations to the classes, we normalize all the pixel values into [-128,127] and add an offset of 128 to create a standard grey scale image. Thus, pixels that are white have a high positive correlation with the target neuron, and pixels which are black have a high negative correlation. Grey pixels are uncorrelated. Figure \ref{fig:am_mnist_pure} (Left) shows the result without using any regularization. This noise is expected due to the lack of control of the image manifold - here we are sampling from pure pixel space, and uncorrelated neurons will have random values. Figure \ref{fig:am_mnist_pure} (Right) shows the result when we enforce L2 regularization - forcing LDAM to sample from the gradient space. As we can see, the results have taken on a new meaning: the dark patches should have very high negative correlation with the output class, while light patches have very high positive correlation.

One sample from the distribution contains relatively little information about the behavior of a neuron. If we instead compute the average of the samples for each sampling procedure, patterns start to emerge, as shown in Figure \ref{fig:am_mnist_avg} (left). The uncorrelated pixels average over time to a gray value, while correlated pixels average to their correlation value. Figure \ref{fig:am_mnist_avg} (Right) gives some examples when we use both L2, and sample averaging.


\subsection{Parameter averaging}
\label{sec:parameter_avg}

As shown in Fig \ref{fig:am_mnist}, by using LDAM with L2 regularization and sample averaging, we are able to find useful diagnostic images corresponding to pixel-gradient correlation. While interesting on its own, the true power of the technique can be shown by exploring how the training process of the classifier can influence how the network responds to different stimuli.

Parameter averaging techniques make up a common set of techniques used to improve classification performance. Parameter averaging works by taking snapshots of the model parameters $\theta$ during the training process, and then averaging them to compute the overall model parameters. Mathematically, we can consider each $\theta_t$ of a model trained using stochastic gradient descent as an estimate of the true model parameters, thus we can use these estimates to compute the expectation of the model parameters $\theta$ by simple averaging. In this experiment we train the LeNet-5 model for ten epochs achieving a 98.7\% accuracy, and for the last five epochs we compute a moving average of the model parameters reaching a final accuracy of 99.11\%. We also computed the average prediction of those model samples in the last five epochs, giving an accuracy of 99.09\%.



\begin{figure}[t]
\centering
\includegraphics[width=\linewidth]{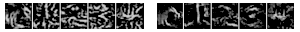}
\caption{Left: Samples from classes 0-4 in a model trained with parameter averaging. Right: Samples from a model trained using normal gradient descent. Both use L2 normalization and sample averaging. Using LDAM we can see multiple modes, while in AM only a single mode is visible. }
\label{fig:param_avg}
\vspace{-4mm}
\end{figure}

We again use LDAM with $L2$ regularization and sample averaging to compute the image samples for the 10 pre-softmax output layer neurons from a model trained using parameter averaging. The resulting samples are shown in Figure \ref{fig:param_avg}. Compared to the base model, we notice a multiple-mode effect in the images generated by the averaged model. For example, in the base model the four has only a single vertical black stripe in the center of the image, while in the averaged model there are many unique vertical black stripes. This implies that the averaged model is activated by a more diverse set of input images.

While the parameter averaged models seem to have a more pronounced multiple mode effect, they also appear to have many different areas of gradient response. Since the L2-normalized images are indications of the gradient, we notice that there are multiple localities that can be activated for the model to positively classify an image in the parameter-averaged model, while the same diversity of responses is not present in the traditional model. This suggests that parameter averaging increases the robustness of a network by improving the robustness to multiple-modalities in the image - a novel insight.


Visualizing the multi-modal nature of the neurons is a clear benefit of using sample-based activation maximization techniques such as LDAM. Because we are sampling directly from the posterior distribution, by using sample averaging we can visualize the multiple modes in the gradient of a neuron easily and efficiently. If we were using classical gradient-based activation, we would only be able to visualize these modes one at a time, and only at random based on the initialization.



\subsection{Adversarial Discrimination}

In the previous section, we showed that using LDAM, $L2$ regularization, and sample averaging to visualize the gradients is a powerful means of exploring the differences in classification performance between two classifiers. In this section we explore the ideas presented in Section \ref{sec:discriminator}, and explore how we can use discriminators to smoothly explore the image manifold.

We train the discriminator at training time, tracking the steps described in Section \ref{sec:discriminator}. At sampling time, we provide a control for the weight of the gradient from the discriminator (the value $\lambda$ in Equation \ref{eq:ldam}). If the weight is set to 0, it becomes the normal LDAM sampling process with no discriminator function. If the weight is very high, the discriminator will overpower the activation neuron, and the sampling technique will focus on sampling only from the discriminator allowed space. Thus, we can explore the boundary between the image manifold and the highly activating images by using a trade-off between discriminator loss and neuron activation loss. 

The results of this method for the MNIST dataset can be seen in Figure \ref{fig:headline} on the first page. In this image, the columns represent the output neurons, while the rows correspond to different weights of the discriminator (Weights from top to bottom: 0, 0.2, 0.5, 0.8.,1.0). In this image we can see that even a slight weight to the discriminator can quickly clean up noise in the image (as by feeding the discriminator real MNIST images, it quickly learns that those images should be sparse). In addition, we also notice that the increasing discriminator weight smoothly trades off between pixel-level interpretability at the lowest discriminator levels, and global image "visual interpretability" with higher discriminator-weight images looking clearly like numbers that could be sampled from MNIST. In the last column of Figure \ref{fig:headline}, we can visualize samples from \textit{only} the learned discriminator image manifold, while ignoring the classifier completely, giving some insight into the kinds of images which our regularizer has learned to generate.

\section{Case Study: CIFAR}
\label{sec:cifar}

\begin{figure}
\centering
\includegraphics[width=\linewidth]{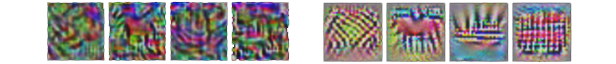}
\caption{Comparing images generated with LDAM (Right) and traditional activation maximization (Left) with only L2 regularization. Clearly LDAM outperforms traditional methods.}
\label{fig:cifar}
\end{figure}

\begin{figure}
\includegraphics[width=\linewidth]{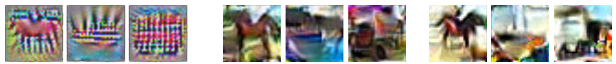}
\caption{Images generated by LDAM for the Horse, Ship, and Truck output neurons from the VGG-16 network, under different discriminator weights. Left: 0.2, Middle: 0.5, Right: 0.8. } 
\label{fig:cifar_vgg_dis}
\vspace{-12pt}
\end{figure}

The MNIST dataset is relatively simple, however most real problems lie in much more complicated spaces. Thus, we move on to exploring models trained on the CIFAR-10 dataset \cite{krizhevsky2009learning} which contains 50000 tiny $32 \times 32$  color images. This dataset has more diverse textures and objects, and presents a more interesting challenge for a visualization technique. In this study we use a VGG net\cite{simonyan2014very} which achieves 85.3\% accuracy on the CIFAR-10 validation set.

We use LDAM to generate image samples activating the output neurons, again selecting neurons before the softmax layer to get more interpretable images. We use both the L2 regularization and sample averaging techniques. The results can be seen in Figure \ref{fig:cifar} which shows that the LDAM method can generate more interpretable images in this case. 

\begin{figure}
\centering
\includegraphics[width=\linewidth]{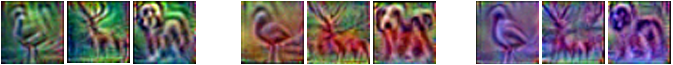}
\caption{Examples of the Bird, Deer and Dog pre-softmax output neurons in CIFAR-10 with the VGG model using L2, averaging, and discrimination, from different points during training. Not only does LDAM sample the class priors, but different global color temperatures.}
\label{fig:color}
\vspace{-3mm}
\end{figure}

We can also apply discrimination to the model, as we did in the MNIST examples. Here we use a basic discriminator with 3 convolution layers and 2 FC layers (Similar to the basic model). Generated samples are given in Figure \ref{fig:cifar_vgg_dis}. We can observe that, the discriminator makes images in the `ship' and `truck' classes more recognizable compared to the original results in Figure \ref{fig:cifar}. Further, we can see that the image samples now contain smoother features, which we would expect in a real-world model. In addition to this, we are still able to explore different features of the image manifold. In Figure \ref{fig:color}, we can see that samples are generated with different global color temperatures, reflecting the model's invariance to overall image average temperature.

\begin{figure}
\centering
\begin{subfigure}[t]{0.9\linewidth}
\centering
\includegraphics[width=\linewidth]{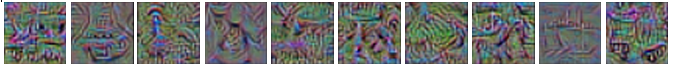}
\caption{From a model trained using normal gradient descent}
\label{fig:cifar_vgg_l2_avg_pavg}
\end{subfigure}
\hfill
\begin{subfigure}[t]{0.9\linewidth}
\centering
\includegraphics[width=\linewidth]{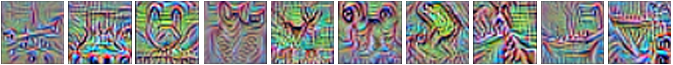}
\caption{From a model trained using parameter averaging}
\label{fig:cifar_model_avg_averaging}
\end{subfigure}
\begin{subfigure}[t]{0.9\linewidth}
\centering
\includegraphics[width=\linewidth]{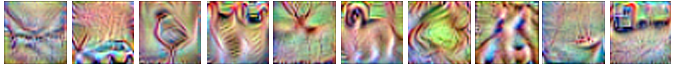}
\caption{From a model trained using parameter averaging and discriminator regularization}
\label{fig:cifar_model_avg_averaging_dis}
\end{subfigure}
\caption{Comparing activation maximization results for the VGG-16 models trained with different methods of parameter averaging. Notice there are multiple modes visible in the parameter-averaged model, particularly in the bird, cat, and dog classes. (a) and (b) use 0 discriminator gradient, with L2 and sample averaging. (c) further adds discriminator gradient. Each column corresponds to a different class (From left to right: Airplane, Car, Bird, Cat, Deer, Dog, Frog, Horse, Boat, Truck)}
\label{fig:cifar_pavg}
\vspace{-10pt}
\end{figure}

In addition, we further demonstrate the multi-modal effect of parameter averaging. We follow the same steps described in section \ref{sec:parameter_avg} to obtain a VGG-16 model trained with parameter averaging. The results are shown in Fig \ref{fig:cifar_model_avg_averaging}. We can see that like in MNIST, the parameter averaged model contains multiple modes for each object, unlike in the original model which contained only single representative feature filters. It is interesting to note that in Figure \ref{fig:cifar_model_avg_averaging_dis} although the discriminator gradient helps improve the image-level feature interpretability, the multi-modal property is degraded due to the discriminator's influence.

The experiments in parameter averaging with MNIST and CIFAR allow us to connect the dots between ensemble learning, which has been well studied, SmoothGrad \cite{smilkov2017smoothgrad} and sample/parameter averaging methods. If we obtain only a single solution from a non-convex optimization problem, no matter whether it is a model or a sample, it could be noisy and imperfect. However, if we introduce noise and variance in the optimization to generate a series of samples, and compute the `average' of them appropriately, we can find improved results, due to the smoothing effects over the multi-modal behavior.

\section{Case Study: ImageNet}
\label{sec:ImageNet}

While experiments on small datasets can be insightful, we would like our methods to be applicable to traditional and modern computer vision problems. To this end, we demonstrate the usage of our algorithm on models trained with the ImageNet dataset. The ImageNet dataset contains more than 1 million 256*256 RGB images in 1000 different classes. Many network architectures such as AlexNet\cite{krizhevsky2012imagenet}, VGG net \cite{simonyan2014very} and ResNet\cite{he2016deep} have been proposed - and we can see that LDAM can illuminate some of the differences between the models. In this case study, we use the AlexNet \cite{krizhevsky2012imagenet} and ResNet \cite{he2016deep} architectures. We train these models on ImageNet from scratch with the RMSProp optimizer. 

\begin{figure}
\centering
\includegraphics[width=0.8\linewidth]{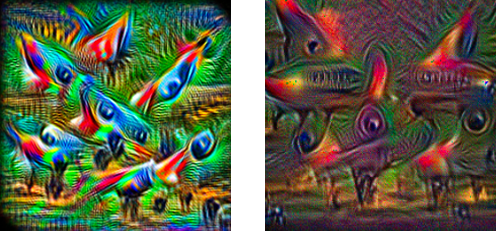}
\caption{Visualization of the Airplane pre-softmax output neuron in ResNet (Left) and AlexNet (Right) using 0.75 discrimination weight, and L2 regularization with averaging.}
\label{fig:rn_v_an}
\vspace{-10pt}
\end{figure}

Figure \ref{fig:rn_v_an} shows the difference under discrimination between the ResNet and AlexNet style architectures. Both of these figures use the same LDAM parameters with 0.75 discrimination weight and L2 $(\lambda=0.1)$ and sample averaging (window of 200 frames). We can see that there are many different receptive fields in the ResNet architecture and that the responses fall into smaller parts of the image than in the larger receptive fields of the AlexNet model. 

\begin{figure}
\centering
\includegraphics[width=0.8\linewidth]{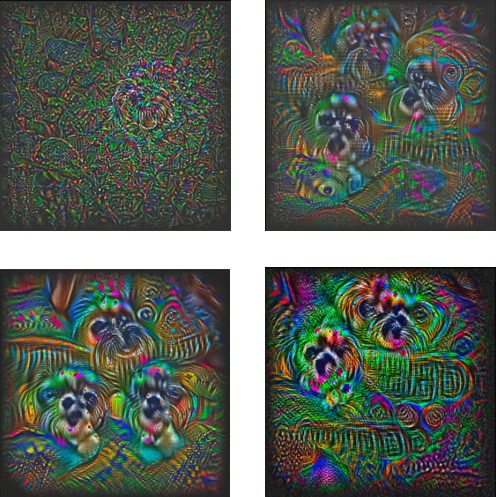}
\caption{The effect of discrimination on the output neuron of the pre-softmax FC8 neuron in the ResNet for the Shih-Tzu class. Top Left: 0.0 discrimination weight, Top Right: 0.25, Bottom Left: 0.50, Bottom Right: 0.75. All of the images use L2 regularization and sample averaging.}
\label{fig:imagenet}
\end{figure}

Figure \ref{fig:imagenet} shows the power of discrimination in the ImageNet space. With very little discrimination, we can see pixel-level features which can help us understand some of the local shapes that the network is responding to. With higher levels of regularization using the discriminator, we can see more global structures. The power of LDAM is to transition at will between these representations in real time with an explicit understanding of how the regularization is affecting the generated visualizations.

\section{Discussion \& Conclusion}
In this paper, we have present LDAM, an SGLD-based monte-carlo sampling algorithm that can generate images activating selected neurons in a deep network. In addition, we have discussed some of the pitfalls of current AM methods, which are tending towards regularization trends which are similar to real-world images, over useful pixel-level diagnostics. We introduce a principled way of exploring regularization and demonstrate the effectiveness of LDAM across three common vision datasets. In addition to demonstrating the multi-modal behavior of LDAM, we also find a novel insight into parameter averaging, which is impossible to visualize with current AM or GAN based techniques. 

While LDAM represents a good first step towards using sampling-based methods, significant future work remains in this area, including the definition of more flexible regularization techniques, and better methods of visualizing and labeling internal neurons in large networks. It is clear, however, that sample-based methods for diagnostic visualization can help to supplement existing end-to-end and GAN based methods in a diagnostician's toolbox. The LDAM code is made is publicly available at \url{https://github.com/BIDData/BIDMach/blob/master/readme_gui.md}.

\section*{Acknowledgements}

We gratefully acknowledge the support of NVIDIA Corporation with the donation of the Titan X GPU used for this research. We additionally acknowledge the support of the Berkeley Artificial Intelligence Research (BAIR) Lab. This work is supported in part by the DARPA XAI program.



{\small
\bibliographystyle{ieee}
\bibliography{egbib}
}

\end{document}